%% file: main.tex
\documentclass[10pt,twocolumn,letterpaper]{article}

\usepackage[pagenumbers]{cvpr} 
\input{preamble}
\definecolor{cvprblue}{rgb}{0.21,0.49,0.74}
\usepackage[pagebackref,breaklinks,colorlinks,allcolors=cvprblue]{hyperref}

\def\paperID{*****} 
\def\confName{CVPR}
\def\confYear{2026}

\title{OmniUMI: Towards Physically Grounded Robot Learning via Human-Aligned Multimodal Interaction}

\author{
Shaqi Luo$^{1}$\thanks{Equal Contribution}, Yuanyuan Li$^{1,2,3}$\footnotemark[1], Youhao Hu$^{1}$\footnotemark[1], Chenhao Yu$^{1,4}$\footnotemark[1],\\ Chaoran Xu$^{1,5}$, Jiachen Zhang$^{1,4}$, Guocai Yao$^{1}$, Tiejun Huang$^{1,6}$, Ran He$^{2,3}$, Zhongyuan Wang$^{1}$\\
$^{1}$Beijing Academy of Artificial Intelligence, \\
$^{2}$MAIS \& NLPR, Institute of Automation, Chinese Academy of Sciences, \\
$^{3}$School of Artificial Intelligence, University of Chinese Academy of Sciences, \\
$^{4}$Beijing Institute of Technology, \\
$^{5}$Beijing University of Posts and Telecommunications, \\
$^{6}$Peking University
}

\begin{document}
\maketitle
\input{sec/0_abstract}
\input{sec/1_intro}
\input{sec/2_related_work}
\input{sec/3_method}
\input{sec/4_experiment}
\input{sec/5_conclusion}
{
    \small
    \bibliographystyle{ieeenat_fullname}
    \bibliography{main}
}


\end{document}

%% file: sec/0_abstract.tex
\begin{figure*}[t]
    \centering
    \includegraphics[width=\textwidth]{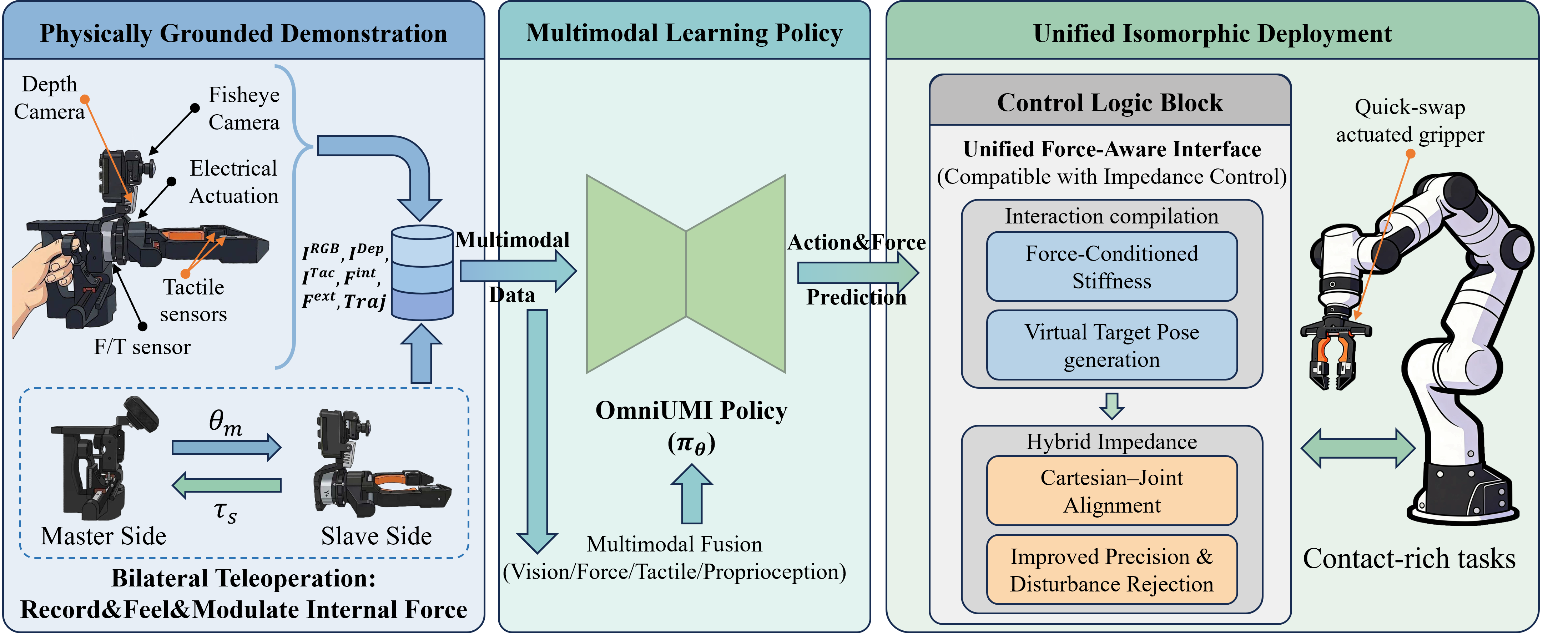}
    \caption{\textbf{OmniUMI overview.}
    \textit{Left:} a unified multimodal handheld interface for physically grounded data acquisition, synchronously capturing RGB, depth, trajectory, tactile sensing, internal grasping force, and external interaction wrench, with bilateral gripper feedback and a handheld embodiment for human-aligned demonstration. 
    \textit{Middle:} multimodal policy learning over visual, tactile, and force-related observations. 
    \textit{Right:} impedance-compatible deployment, where policy outputs are translated into virtual targets and executed for unified regulation of motion and contact behavior in contact-rich manipulation.}
    \label{fig:teaser}
\end{figure*}

\begin{abstract}

UMI-style interfaces enable scalable robot learning, but existing systems remain largely visuomotor, relying primarily on RGB observations and trajectory while providing only limited access to physical interaction signals. This becomes a fundamental limitation in contact-rich manipulation, where success depends on contact dynamics such as tactile interaction, internal grasping force, and external interaction wrench that are difficult to infer from vision alone. We present OmniUMI, a unified framework for physically grounded robot learning via human-aligned multimodal interaction.\footnote{Project page: \url{https://baai-aether.github.io/OmniUMI}} OmniUMI synchronously captures RGB, depth, trajectory, tactile sensing, internal grasping force, and external interaction wrench within a compact handheld system, while maintaining collection--deployment consistency through a shared embodiment design. 
To support human-aligned demonstration, OmniUMI enables natural perception and modulation of internal grasping force, external interaction wrench, and tactile interaction through bilateral gripper feedback and the handheld embodiment.
Built on this interface, we extend diffusion policy with visual, tactile, and force-related observations, and deploy the learned policy through impedance-based execution for unified regulation of motion and contact behavior. Experiments demonstrate reliable sensing and strong downstream performance on force-sensitive pick-and-place, interactive surface erasing, and tactile-informed selective release. Overall, OmniUMI combines physically grounded multimodal data acquisition with human-aligned interaction, providing a scalable foundation for learning contact-rich manipulation.

\end{abstract}

%% file: sec/1_intro.tex
\section{Introduction}

Recent advances in robot learning have driven substantial progress in imitation learning and vision-language-action models~\cite{chi_diffusion_2024,rt12022arxiv,kim_openvla_2024,team_octo_2024,liu_rdt-1b_2024,hu_data_2025}. 
At the same time, robot-free data collection has emerged as a practical paradigm for scaling real-world robot learning, as it decouples demonstration acquisition from robot hardware and thereby reduces setup cost, avoids robot wear, and enables scalable data collection by non-expert users~\cite{chi_universal_2024,xu_exumi_nodate,xu_dexumi_2025,choi_--wild_2026}. 
Systems such as UMI (Universal Manipulation Interface)~\cite{chi_universal_2024} and its variants have shown that handheld interfaces can support scalable, portable, and embodiment-agnostic data acquisition. 
Their success suggests that scalable robot learning need not depend on expensive robot-in-the-loop collection.

However, the current evolution of UMI-like systems remains largely confined to visuomotor data. 
Most existing systems rely primarily on RGB observations and trajectory, which are effective for geometric manipulation but provide only limited access to the physical variables that govern contact-rich behavior. 
Yet many real-world manipulation skills depend critically on contact dynamics, including tactile sensing, internal grasping force, and external interaction wrench, which are difficult to infer reliably from vision alone. 
As a result, a broad class of tasks---including wiping, screwing, deformable object handling, compliant assembly, and fragile grasping---remains difficult to learn from vision-dominant data alone.

A natural next step is therefore to extend robot-free interfaces from visuomotor data collection toward richer interaction-centric data~\cite{wu_tacdiffusion_2024,zhou_admittance_2024,yang_moma-force_2023,adeniji_feel_2025}. 
Recent efforts have explored augmenting UMI-style systems with tactile sensing~\cite{liu_vitamin_2025,xu_exumi_nodate} or force/torque sensing~\cite{yu_forcevla_2025,lee_manipforce_2025,liu_forcemimic_2024}, showing that richer physical feedback can improve performance. 
Nevertheless, these approaches are typically modality-centric: they validate individual sensing channels in isolation, but do not address the broader challenge of building a unified framework for \emph{physically grounded robot learning via human-aligned multimodal interaction}. 
For contact-rich manipulation, it is not sufficient to merely add more sensing channels. 
The collected data should be \emph{physically grounded}, in the sense that they directly capture physically meaningful interaction variables such as tactile sensing, internal grasping force, and external interaction wrench. 
At the same time, the demonstration process should be \emph{human-aligned}, in the sense that the operator can naturally perceive and regulate these variables during data collection so that the recorded signals remain aligned with human intent.

This challenge is fundamentally a systems problem. 
Integrating tactile sensing requires careful finger design, sensor protection, and signal representation, often at the cost of reduced manipulation capability or increased fragility. 
Incorporating force/torque sensing introduces additional complexity in wiring, calibration, parameter identification, gravity compensation, and coordinate transformation from the sensor frame to the end-effector frame. 
Moreover, mechanical coupling between actuation and sensing can directly contaminate force measurements~\cite{liu_forcemimic_2024,lee_manipforce_2025}, making accurate multimodal acquisition especially difficult in handheld settings. 
More broadly, when multiple heterogeneous sensing modalities must coexist within a single compact interface, additional challenges arise, including sensing interference, fragile hardware exposure, asynchronous communication, and mismatch between data-collection and deployment embodiments.

In this work, we present \textbf{OmniUMI}, a unified framework for \emph{physically grounded robot learning via human-aligned multimodal interaction}. 
Rather than treating tactile, force, and motion signals as independent add-ons, OmniUMI is designed around the concept of \emph{interaction acquisition}, in which policies are trained directly on physically meaningful multimodal interaction signals collected through a human-aligned demonstration interface.
To realize this goal, our framework is built around three tightly coupled components: 
(i) a unified multimodal handheld interface for physically grounded data acquisition, 
(ii) a human-aligned demonstration pipeline centered on multimodal physical interaction, and 
(iii) a multimodal learning and impedance-compatible deployment framework. 
Together, these components establish a consistent pipeline from multimodal human interaction to contact-rich robot execution.

At the system level, OmniUMI synchronously captures RGB, depth, trajectory, tactile sensing, internal grasping force, and external interaction wrench within a compact handheld system, while maintaining collection--deployment consistency through a shared embodiment design. 
A custom hardware hub unifies communication, protocol conversion, voltage regulation, and device management through a single connection, enabling practical multimodal integration. 
A motorized quick-swap gripper is reused across both data collection and execution, reducing embodiment mismatch and ensuring sensing consistency. 
A biomimetic finger design further enables protected tactile integration without sacrificing manipulation capability. 
By jointly acquiring vision, motion, tactile, and force-related signals within one reusable interface, OmniUMI moves beyond visuomotor supervision toward physically grounded multimodal interaction data.

At the demonstration level, OmniUMI emphasizes \emph{human-aligned interaction}. 
To better align recorded demonstrations with operator intent during contact-rich manipulation, OmniUMI grounds multimodal physical interaction through bilateral gripper feedback and the handheld embodiment: internal grasping force is rendered through a custom master--slave gripper with bilateral control, while the handheld form factor naturally preserves perception of external interaction wrench and tactile interaction.
Together, these mechanisms allow operators to naturally perceive, modulate, and record internal grasping force, external interaction wrench, and tactile interaction during demonstration.
In this sense, OmniUMI does not merely measure physical variables; it organizes their acquisition around a human-aligned interface for natural interaction regulation.

At the learning and deployment level, we extend diffusion policy (DP)~\cite{chi_diffusion_2024} to incorporate multimodal observations, including visual, tactile, and force-related signals, enabling policies to reason jointly over geometry and interaction. 
On the deployment side, we translate policy outputs into virtual targets and execute them via impedance-based control, enabling implicit and unified regulation of motion and force without explicit switching between control modes. 
This yields a consistent pipeline from physically grounded, human-aligned data acquisition to contact-rich policy execution.

We evaluate OmniUMI on three representative manipulation settings: 
(1) force-sensitive pick-and-place, 
(2) interactive surface erasing, and 
(3) tactile-informed selective release. 
These tasks directly correspond to the three key interaction capabilities emphasized in our framework: grasp-force-aware manipulation, wrench-informed surface interaction, and tactile-informed fine-grained control. 
Experimental results show that OmniUMI enables more robust and physically consistent learning in contact-rich scenarios, while reducing operator burden and narrowing the gap between robot-free data collection and real-world deployment.

\noindent\textbf{Contributions.}
The contributions of this work are three-fold:
\begin{itemize}
    \item \textbf{A unified multimodal robot-free interface for physically grounded data acquisition.}
    We introduce a compact handheld system that jointly captures RGB, depth, trajectory, tactile sensing, internal grasping force, and external interaction wrench, moving beyond visuomotor data toward physically grounded multimodal interaction acquisition.
    
    \item \textbf{A human-aligned multimodal interaction acquisition paradigm via bilateral gripper feedback and handheld embodiment.}
    We introduce a demonstration interface in which multimodal physical interaction is grounded through bilateral gripper feedback and the handheld embodiment: internal grasping force is rendered through bilateral control in a custom master--slave gripper, while external interaction wrench and tactile interaction are naturally grounded through the handheld embodiment, enabling these interaction signals to be perceived, modulated, and recorded during demonstration and improving alignment between recorded interaction signals and human intent.

    \item \textbf{A consistent multimodal learning and impedance-based deployment pipeline.}
    We integrate visual and interaction signals into a unified policy framework and deploy them via impedance-based control, enabling stable and implicit regulation of motion and force without controller switching.
\end{itemize}

By combining physically grounded multimodal data acquisition with human-aligned interaction, OmniUMI provides a scalable foundation for learning contact-rich manipulation beyond vision-only representations.

%% file: sec/2_related_work.tex
\section{Related Work}

\subsection{Robot-free Interfaces for Scalable Robot Learning}

Robot-free data collection has emerged as a practical paradigm for scaling real-world robot learning, as it decouples demonstration acquisition from robot hardware and avoids the cost, wear, and operational constraints associated with robot-in-the-loop data collection. 
Handheld interfaces such as UMI~\cite{chi_universal_2024} established a representative pipeline for portable robot-free data collection using RGB observations, relative motion reconstruction, and gripper state, enabling embodiment-agnostic imitation learning. 
Follow-up systems such as FastUMI~\cite{zhaxizhuoma_fastumi_2025} and DexWild~\cite{tao_dexwild_2025} further improved scalability, hardware independence, and ease of deployment. 
More recently, large-scale studies such as RDT2~\cite{liu2026rdt2}, along with industrial efforts including Generalist AI and Sunday Robotics, have demonstrated that robot-free data pipelines can support cross-embodiment generalization and large-scale robot learning.

Despite these advances, most robot-free systems remain primarily visuomotor-centric, with RGB observations and trajectories serving as the dominant supervision signals. 
While such representations are effective for free-space motion and simple pick-and-place tasks, they provide limited information about contact dynamics, force regulation, and other physical variables that are critical for contact-rich manipulation. 
From the perspective of our work, this marks the central limitation of current robot-free interfaces: they have demonstrated scalability, but their data representations remain largely insufficient for physically grounded interaction learning.

\subsection{From Visuomotor Interfaces to Physically Grounded Multimodal Interaction}

A growing body of work has begun extending robot-free interfaces beyond pure visuomotor supervision by incorporating richer physical modalities~\cite{wu_tacdiffusion_2024,zhou_admittance_2024,yang_moma-force_2023,adeniji_feel_2025}. 
In particular, tactile sensing~\cite{liu_vitamin_2025,xu_exumi_nodate}, internal grasping force~\cite{choi_--wild_2026}, and external interaction wrench~\cite{liu_forcemimic_2024,lee_manipforce_2025} have all been shown to provide important information that is difficult to infer reliably from vision alone. 
These developments suggest a broader shift: if robot-free learning is to extend from geometric manipulation to contact-rich manipulation, the key challenge is no longer only scalable data collection, but scalable acquisition of \emph{physically grounded} multimodal interaction data.

However, prior efforts are still largely modality-centric. 
Most systems augment existing pipelines with one additional sensing channel at a time, validating the usefulness of that modality in isolation, but without fully addressing how multiple heterogeneous interaction variables should be jointly integrated into a compact, reusable, and deployment-compatible robot-free interface. 
This leaves open the broader systems question that motivates our work: how to move from scalable visuomotor interfaces to scalable \emph{physically grounded robot learning via human-aligned multimodal interaction}, in which tactile interaction, internal grasping force, and external interaction wrench are not only measured, but also naturally perceived and regulated during demonstration.

\subsection{Tactile Sensing in Robot-free Manipulation Interfaces}

To address the limitations of purely visual observations, a growing body of work incorporates tactile sensing into robot-free interfaces. 
Vision-based tactile sensors and fingertip tactile arrays provide localized information about contact geometry, friction, slip, and deformation that cannot be reliably inferred from vision alone. 
Recent hardware advances, including optical, capacitive, and resistive tactile sensors~\cite{bhirangi_anyskin_2024}, together with vision-based tactile sensing approaches such as GelSight~\cite{helmut_tactile-conditioned_2025} and DIGIT~\cite{liang_alltact_2025}, have significantly improved the practicality of tactile perception. 
Systems such as Vitamin~\cite{liu_vitamin_2025}, TacUMI~\cite{cheng_tacumi_2026}, and TacThru-UMI~\cite{li_simultaneous_2025} demonstrate that tactile sensing substantially improves performance in contact-rich manipulation, particularly for slip detection and fine contact adaptation.

However, the key challenge is no longer the utility of tactile sensing itself, but how to incorporate it into a practical robot-free interface. 
Many existing designs integrate tactile sensors directly into the gripper fingertips~\cite{cheng_tacumi_2026}, which is mechanically simple but compromises gripper usability, limits small-object manipulation, and exposes sensors to damage. 
At the same time, tactile signals are often high-dimensional and difficult to integrate into policy learning without careful representation design~\cite{feng_anytouch_2026}. 
Therefore, while prior work establishes the importance of tactile sensing, it does not fully resolve how to embed tactile perception into a robust, reusable, and scalable interaction interface that also preserves natural and human-aligned tactile interaction during demonstration.

\subsection{Internal and External Force Acquisition}

Force-related sensing provides another critical modality for contact-rich manipulation and can be broadly categorized into internal force and external force. 
Internal force corresponds to the grasping force applied by the gripper, while external force corresponds to the interaction wrench between the end-effector and the environment. 
Both are essential for physically grounded manipulation, yet prior work typically treats them separately rather than as part of a unified interaction acquisition framework.

For internal force, most robot learning systems do not explicitly measure it, instead relying on gripper position or aperture as a proxy. 
This results in weak representation of grasp regulation and limits the ability to learn force-sensitive behaviors. 
Some works, such as FeelTheForce~\cite{adeniji_feel_2025}, directly capture contact-force information using human-mounted sensing, demonstrating its importance for fragile-object manipulation. 
UMI-FT~\cite{choi_--wild_2026} further explores internal-force acquisition in a robot-free setting. 
However, fingertip-mounted sensors are often fragile and difficult to scale for long-term data collection. 
These limitations motivate more robust approaches to internal-force estimation, such as leveraging motor current and transmission dynamics to provide a scalable and physically grounded representation.

For external force, bilateral teleoperation has long been used to collect high-quality force-aware demonstrations~\cite{luo_humanrobot_2022}. 
Recent robot-free systems, including ForceCapture~\cite{liu_forcemimic_2024}, ManipForce~\cite{lee_manipforce_2025}, and TacUMI~\cite{cheng_tacumi_2026}, demonstrate that interaction wrench significantly improves performance in contact-dominant tasks such as wiping, insertion, and surface interaction. 
However, integrating force--torque sensing into compact handheld interfaces introduces substantial challenges, including sensing interference from mechanically triggered grippers, wiring complexity, calibration, and gravity compensation. 
In particular, trigger-based gripper actuation can contaminate force measurements~\cite{cheng_tacumi_2026}, making reliable external-force acquisition difficult in conventional designs.

Taken together, prior work clearly demonstrates the importance of both internal and external force, but most systems either treat them independently or rely on sensing embodiments that are difficult to scale. 
As a result, the literature still lacks a practical robot-free interface that jointly supports internal force, external force, and complementary modalities within a single reusable system, while also preserving natural and intuitive human interaction during demonstration.

\subsection{Multimodal Policy Learning and Force-aware Deployment}

Beyond data acquisition, multimodal robot learning requires that physical signals such as tactile and force be incorporated into policy representations and coupled effectively with the controller during deployment. 
Recent visuomotor policy learning methods, including diffusion-policy-based approaches~\cite{chi_diffusion_2024,noauthor_reactive_nodate}, have shown strong performance in visually guided manipulation. 
Meanwhile, force-aware and tactile-aware learning methods demonstrate that additional physical modalities can significantly improve performance in contact-rich tasks. 
However, in many systems, these modalities are incorporated only as auxiliary inputs, and deployment still relies on controller switching or rule-based adjustments rather than a unified control interface.

Some recent approaches attempt to bridge this gap by compiling force-related policy outputs into virtual target poses and compliance parameters, enabling implicit force regulation. 
For example, ACP-style pipelines~\cite{hou_adaptive_2024} and UMI-FT~\cite{choi_--wild_2026} map learned force signals to virtual targets for compliant execution. 
However, these approaches are primarily validated in admittance-style control, and extending them to impedance-based execution remains challenging due to limitations in stiffness tuning and stability. 
This is particularly important since many real robotic systems, especially humanoids, rely on impedance-compatible control interfaces. 
These limitations suggest that a complete solution requires not only multimodal data collection but also a unified learning-and-deployment framework that is compatible with practical control architectures.

\subsection{Summary and Our Position}

In summary, prior work shows a clear progression from scalable robot-free visuomotor data collection~\cite{chi_universal_2024,zhaxizhuoma_fastumi_2025,tao_dexwild_2025}, to tactile-augmented manipulation~\cite{liu_vitamin_2025,bhirangi_anyskin_2024,liang_alltact_2025}, to force-aware interaction learning~\cite{luo_humanrobot_2022,liu_forcemimic_2024,lee_manipforce_2025,adeniji_feel_2025}, and further toward multimodal policy learning and force-aware deployment~\cite{chi_diffusion_2024,xue2025reactive}. 
However, most existing systems extend robot-free interfaces by adding one sensing modality at a time, without fully addressing the deeper transition from scalable visuomotor data collection to \emph{physically grounded} and \emph{human-aligned} multimodal interaction.

Our work is positioned beyond modality-specific extensions. 
Rather than introducing only an additional sensing channel, we aim to build a unified robot-free framework that jointly supports RGB, depth, trajectory, tactile sensing, internal grasping force, and external interaction wrench within a single reusable system. 
Moreover, we emphasize both \emph{physically grounded} multimodal data acquisition and \emph{human-aligned} interaction acquisition: the former by explicitly capturing physically meaningful interaction variables, and the latter by organizing their collection around natural human perception and regulation of internal grasping force, external interaction wrench, and tactile interaction through bilateral gripper feedback and the handheld embodiment.
Combined with controller-compatible multimodal policy learning and impedance-based deployment, our system is not merely a tactile- or force-augmented variant of existing robot-free interfaces, but a scalable framework for physically grounded robot learning via human-aligned multimodal interaction.

%% file: sec/3_method.tex
\section{Method}

\subsection{Overview}

We present \textit{OmniUMI}, a unified framework for \emph{physically grounded robot learning via human-aligned multimodal interaction}. 
Unlike prior robot-free systems that primarily collect visuomotor demonstrations, OmniUMI explicitly acquires physically meaningful interaction variables---including tactile sensing, internal grasping force, and external interaction wrench---together with visual observations and motion. 
At the same time, OmniUMI organizes the demonstration process so that these variables are not only measured, but also naturally perceived and regulated by the operator during data collection.

Our approach is built around three tightly coupled components: 
(i) a unified multimodal handheld interface for physically grounded data acquisition, 
(ii) a human-aligned multimodal interaction acquisition pipeline, and 
(iii) a multimodal learning and impedance-compatible deployment framework. 
Together, these components establish a consistent pipeline from robot-free multimodal human interaction to contact-rich robot execution.

Rather than treating tactile, force, and motion signals as independent add-ons, OmniUMI is designed around the concept of \emph{interaction acquisition}, where policies are trained directly on physically meaningful multimodal interaction signals.
This perspective shapes not only the sensing embodiment, but also the structure of the collected data, the semantics of the learned policy interface, and the way learned actions are executed on the robot.

\subsection{Design Objectives}

OmniUMI is designed as a scalable robot-free framework for physically grounded robot learning via human-aligned multimodal interaction. 
Rather than incrementally attaching sensing modules to an existing hardware form, we start from system-level objectives that jointly shape the hardware embodiment, data acquisition process, and downstream learning-and-deployment pipeline.

\textbf{Physically Grounded Multimodal Data Acquisition.}
A practical robot-free interface for contact-rich manipulation should not rely only on RGB observations and motion trajectories. 
Instead, it should explicitly acquire physically meaningful interaction variables, including tactile sensing, internal grasping force, and external interaction wrench, so that the collected data better reflect real contact dynamics. 
A key design goal is therefore to integrate RGB, depth, trajectory, tactile sensing, internal force, and external force within one compact system despite substantial differences in communication protocol, power requirement, update rate, and mechanical coupling.

\textbf{Human-Aligned Interaction Acquisition.}
Collecting more sensing channels is not sufficient if the resulting data do not faithfully reflect how humans regulate interaction. 
OmniUMI therefore aims to acquire physical variables through an interface that allows the operator to naturally perceive and modulate internal grasping force, external interaction wrench, and tactile interaction during demonstration. 
This objective motivates our co-design of bilateral gripper feedback and the handheld embodiment, through which these interaction variables remain aligned with the operator's natural contact experience during data collection.

\textbf{Collection--Deployment Consistency.}
In typical UMI-style pipelines, data collection and deployment are often performed with two different embodiments, requiring additional effort to align robot-side observations with those seen during demonstration. 
This issue is already nontrivial when observations mainly consist of RGB and gripper width. 
Once tactile sensing, internal force, and external force are introduced, however, the mismatch becomes substantially more severe, because these multimodal observations depend not only on scene appearance but also on the sensing embodiment itself and the underlying contact mechanics. 
OmniUMI therefore aims to minimize hardware mismatch between data collection and deployment so that multimodal observations remain as consistent as possible across both stages.

\textbf{Unified Controller-Compatible Deployment.}
Multimodal interaction learning is only useful if the learned outputs can be realized under the controller constraints of real robots. 
In contact-rich manipulation, this requirement goes beyond compliance alone: the deployment framework should not be limited to admittance-based execution, but must also remain compatible with Cartesian-space impedance control, which is widely used in practical robot systems and is especially relevant to humanoids. 
Moreover, force-aware deployment should avoid explicit switching between position-control and force-control modes, since such switching often introduces discontinuity, tuning complexity, and reduced robustness under changing contact conditions. 
OmniUMI therefore treats unified controller-compatible deployment as a design objective from the outset, so that multimodal interaction outputs can be translated into stable and physically meaningful robot behavior within a single control framework.

\subsection{System Architecture}

To realize the objectives above, OmniUMI is built as a coherent handheld system rather than a collection of loosely coupled modules. 
The system design focuses on three questions: how to integrate heterogeneous sensing and actuation into one interface, how to preserve collection--deployment consistency after introducing physically grounded multimodal observations, and how to support human-aligned interaction without sacrificing the practical functionality of the gripper.

\begin{figure*}[t]
    \centering
    \includegraphics[width=\textwidth]{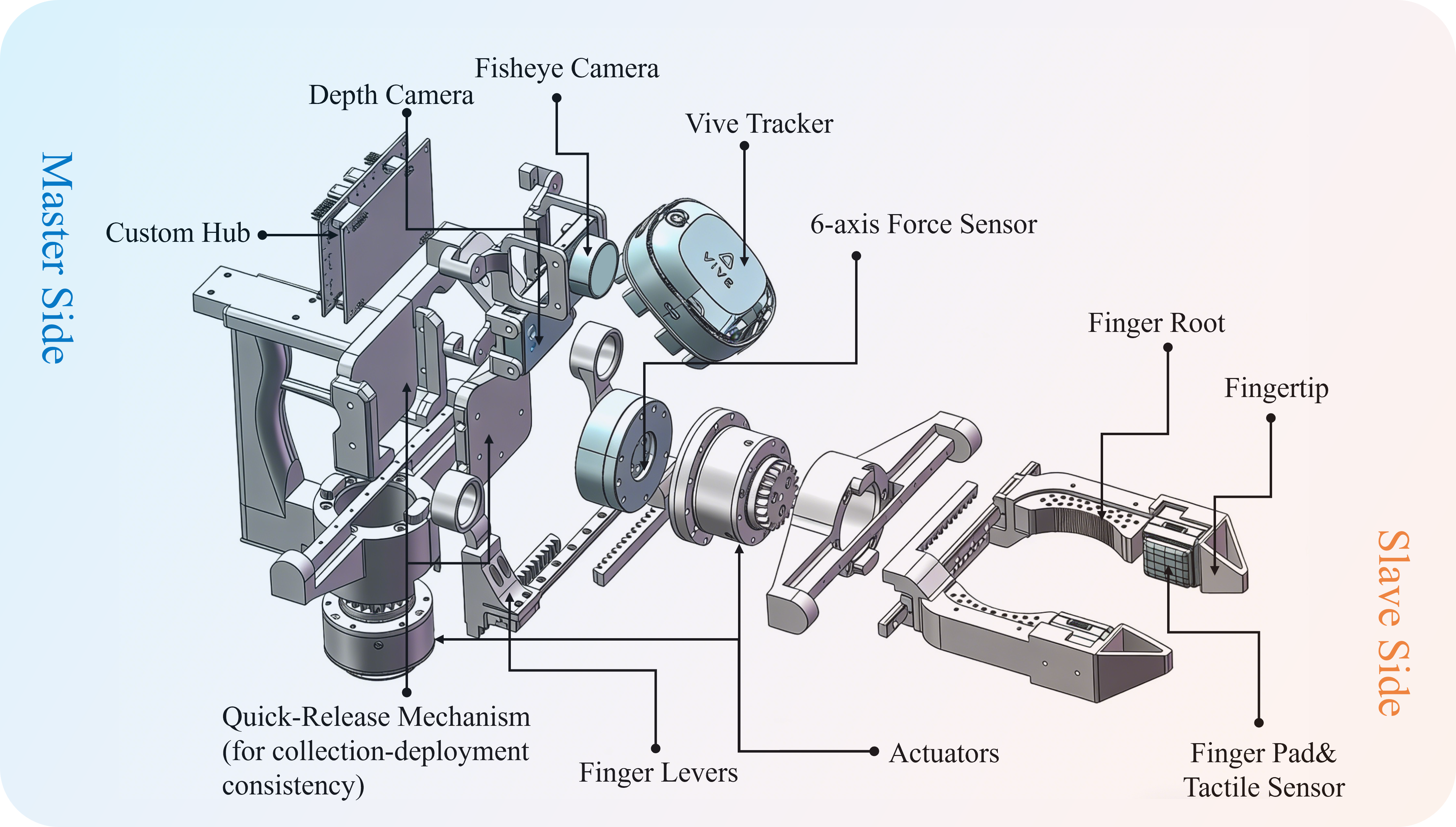}
    \caption{Hardware overview of \textit{OmniUMI}. The system is organized around a unified handheld embodiment that integrates the custom hub, fisheye and depth sensing, 6-axis force sensing, motion tracking, quick-release mechanism, and tactile-compatible gripper structure. The same gripper-side embodiment is reused across data collection and deployment to improve collection--deployment consistency.}
    \label{fig:omniumi_hardware}
\end{figure*}

\paragraph{Multimodal Integration via a Custom Hub.}
A central system challenge in OmniUMI is that all sensing and actuation modules must coexist within one compact handheld device while remaining manageable from the host computer. 
Naive integration would require fragmented wiring, heterogeneous interface logic, and multiple independent power and communication paths, making the system fragile and difficult to use and scale. 
To address this challenge, we build OmniUMI around a custom hub that serves as the central integration layer between the handheld interface and the host computer.

The hub unifies communication, protocol conversion, voltage regulation, and device management for all heterogeneous sensing and actuation modules through a single cable connection. 
This design is particularly important in our setting because the system simultaneously incorporates visual sensing, force-related sensing, tactile sensing, and motorized actuation, all of which would otherwise require separate integration logic. 
By introducing the custom hub as a shared integration layer, OmniUMI turns physically grounded multimodal sensing from a fragmented hardware configuration into a manageable system-level interface.

Beyond reducing wiring complexity, the hub also improves extensibility. 
Since sensing requirements may vary across tasks or future iterations of the system, the interface should support the addition of new modules without requiring a redesign of the entire communication and power architecture. 
The hub therefore serves not only as a hardware connector, but as the system foundation that makes scalable multimodal integration practical.

\paragraph{Collection--Deployment Consistency via a Motorized Gripper.}
A major challenge in UMI-style systems lies in collection--deployment consistency. 
Even in vision-centric settings, aligning the interface used for demonstration with the hardware used for deployment is already nontrivial. 
Once internal and external force, tactile sensing, and depth are introduced, this mismatch becomes substantially more severe because the observations no longer depend only on RGB appearance and gripper width, but also on contact mechanics and sensing embodiment.

To reduce this gap, OmniUMI adopts a motorized quick-swap gripper that can be reused across both collection and deployment. 
Reusing the same gripper preserves not only the geometry of the end-effector, but also the interaction-related sensing context attached to it. 
This significantly reduces alignment effort and makes multimodal observations more consistent across the two stages. 
In this sense, the motorized gripper is not merely a hardware convenience; it is a system design choice aimed at preserving physically grounded multimodal consistency.

The motorized gripper also serves a second purpose: it replaces mechanically triggered actuation with electrical command transmission. 
This becomes critical once force sensing is integrated into the same interface, because mechanically triggered actuation can perturb force measurements and degrade signal quality. 
By shifting actuation to a motorized mechanism, OmniUMI improves both collection--deployment consistency and force measurement cleanliness within the same system design.

\paragraph{Biomimetic Finger Design with Integrated Tactile Sensing.}
Tactile sensing is only useful if it can be integrated without compromising the practical functionality of the gripper. 
In many existing systems, tactile sensors are attached directly to the fingertip region, which often reduces the ability to manipulate small objects and leaves the sensors highly exposed to damage. 
This creates a trade-off between local tactile perception and practical usability.

To mitigate this trade-off, OmniUMI adopts a biomimetic finger design with integrated tactile sensing. 
Rather than treating tactile sensing as a standalone add-on, the finger embodiment is designed so that tactile sensing is incorporated in a more protected and task-compatible manner. 
This allows the interface to preserve the essential functionality of a conventional gripper while benefiting from local tactile perception.

While the present tactile embodiment is not intended as a standalone contribution in tactile sensor design, it serves as a practical system-level solution that makes tactile sensing more usable within a reusable multimodal interaction interface. 
From this perspective, the biomimetic finger design is introduced not as an isolated tactile innovation, but as an embodiment choice that supports durability, fine-object manipulation, and repeated use across collection and deployment.

\subsection{Physically Grounded and Human-Aligned Interaction Acquisition}

With the system embodiment established, the next step is to acquire data that are both physically grounded and human-aligned. 
In OmniUMI, this means that recorded multimodal signals should capture physically meaningful robot--environment interaction variables, while remaining consistent with how the operator perceives and regulates contact during demonstration. 
This requirement is especially important in contact-rich manipulation, where task success depends not only on motion geometry, but also on how grasp force, interaction wrench, and local contact evolve throughout the manipulation process. 
For this reason, OmniUMI organizes data acquisition around three human-aligned physical interaction channels---internal grasping force, external interaction wrench, and tactile interaction---while using RGB, depth, and trajectory as complementary supporting observations.

\paragraph{Internal Force Acquisition with Human-Aligned Bilateral Feedback.}
Internal force plays a central role in manipulation tasks involving fragile, deformable, or compliance-sensitive objects. 
In such settings, successful behavior depends not only on whether the gripper reaches the object, but also on how strongly the object is grasped and how that grasp is modulated during interaction. 
However, in most existing robot learning pipelines, internal force is only weakly represented: the gripper is typically controlled through position, and in many practical systems it is further simplified into a binary open/close command. 
As a result, internal force is typically overlooked rather than explicitly represented in collected demonstrations, making it difficult for learned policies to recover how humans regulate grasp strength in contact-rich tasks.

A straightforward alternative is to rely on tactile sensing to infer grasp interaction. 
Indeed, tactile observations and internal force are closely related, since both reflect contact state, object deformation, and grasp stability. 
However, tactile sensing often remains expensive and fragile, and the representations produced by different tactile sensors can vary substantially across sensing principles, embodiments, and signal formats, making cross-system reuse difficult in practice. 
Internal force therefore provides a complementary and often more practical physical modality: while it does not replace tactile sensing, it offers a comparatively direct, low-dimensional, and mechanically grounded representation of grasp interaction.

To make internal force explicit in OmniUMI, we combine a backdrivable motorized gripper with bilateral feedback. 
This design brings three advantages. 
First, it makes internal force directly measurable from the actuation side through motor current or torque. 
Second, it allows the operator to directly feel and modulate grasp interaction during demonstration, thereby aligning the recorded internal-force signal with human feedback and intent. 
Third, because gripper actuation is transmitted electrically rather than through a mechanically triggered structure, the same design also avoids contaminating the force/torque sensor used for external-force acquisition.

On the measurement side, the grasping force can be estimated from motor-side actuation through a simple transmission model. 
Let \(i_t\) denote the motor current, \(k_\tau\) the motor torque constant, and \(r_g\) the effective meshing radius of the gripper transmission. 
Then the internal grasping force can be approximated as
\begin{equation}
    \tau^g_t \approx k_\tau i_t,
    \quad
    F^{int}_t \approx \frac{\tau^g_t}{r_g}
    = \frac{k_\tau i_t}{r_g}
    \label{eq:internal_force_estimation}
\end{equation}
This formulation provides a compact and physically interpretable estimate of grasp force without introducing additional fingertip force sensors, which would otherwise increase fragility and integration complexity.

On top of this measurement mechanism, OmniUMI uses bilateral feedback to close the loop between grasp command and grasp perception. 
A compact bilateral formulation is given by
\begin{equation}
    \tau_s = K_p \left(b\,\theta_m - \theta_s - \delta\right)
           + K_d \left(b\,\dot{\theta}_m - \dot{\theta}_s\right)
    \label{eq:slave_torque_main_revised}
\end{equation}
\begin{equation}
    \tau_m = -\frac{1}{a}\,\tilde{\tau}_s + B_l\,\dot{\theta}_m
    \label{eq:master_torque_main_revised}
\end{equation}
where $\theta_m,\theta_s$ and $\dot{\theta}_m,\dot{\theta}_s$ denote the motor position and velocity of the master and slave grippers, respectively, $\delta$ is the installation offset between the master and slave sides, $K_p$ and $K_d$ are feedback gains, $b$ is a position scaling factor, $\tilde{\tau}_s$ is the filtered measured torque at the slave side, $a$ controls the reflected torque magnitude, and $B_l$ is a local velocity-feedback term on the master side. 
Under the present sign convention, this term acts not as additional damping injection, but as a compensation of the effective resistance felt at the master side. 
In practice, it offsets part of the sluggishness introduced by torque reflection, thereby reducing the viscous drag perceived by the operator and making master-side manipulation lighter and smoother. 
Through this closed interaction loop, the operator can feel contact onset, object resistance, and grasp transitions directly at the master side rather than inferring them only from visual cues.

This bilateral design improves more than user experience. 
It directly improves the quality of the collected data by making the internal-force signal not only measurable, but also human-aligned. 
In this sense, internal force in OmniUMI is not merely sensed; it is acquired through an interaction loop in which the human can both generate and regulate the variable being recorded. 
More specifically, the slave side realizes PD-based position tracking of the master-side grasp motion, while the master side renders the reflected torque induced by contact at the slave side, allowing the operator to directly perceive grasp interaction during demonstration. 
The local velocity-feedback term further improves operability by compensating part of the reflected impedance, rather than by introducing extra damping. 
At the same time, because the gripper is motorized rather than mechanically triggered, the design also avoids introducing actuation-induced disturbances into the external force/torque sensor. 
As a result, the same hardware choice supports both physically grounded internal-force acquisition and human-aligned interaction during demonstration.

More generally, this design illustrates that interaction variables become human-aligned not only when they are sensed, but when the embodiment allows operators to naturally perceive and regulate them during demonstration.

\paragraph{External Wrench Acquisition and Embodied Perception.}
External force captures the interaction wrench exchanged between the tool and the environment and is therefore essential for contact-dominant tasks such as wiping, scraping, insertion, and surface following. 
In these tasks, successful execution depends not only on where the end-effector moves, but also on how contact is established, maintained, and modulated. 
Compared with purely kinematic observations, external force provides a more direct description of real interaction dynamics, including contact onset, sliding, resistance, and force variation along the task trajectory.

Despite its importance, acquiring clean wrench measurements in a compact handheld system is nontrivial. 
A first challenge is sensing itself: external-force acquisition typically requires additional hardware, calibration, and compensation. 
A second and often more subtle challenge is contamination introduced by the interface mechanism. 
In many handheld UMI-like systems, gripper actuation is mechanically triggered, so the actuation process itself perturbs the force/torque sensor and contaminates the recorded wrench. 
In this case, the measured signal no longer reflects only robot--environment interaction, but also artifacts introduced by the interface embodiment.

OmniUMI addresses this issue by combining compact force/torque sensing with motorized electrical actuation. 
Since gripper commands are transmitted electrically to a motorized gripper rather than through a mechanically triggered linkage, actuation-induced contamination is eliminated at the source. 
To ensure coordinate-consistent and bias-reduced wrench estimation, we compute the compensated wrench as
\begin{equation}
    \mathbf{w}^{ee}_t
    =
    \mathcal{T}_{s\rightarrow ee}
    \left(
        \mathbf{w}^{raw}_t - \mathbf{w}^{grav}_t
    \right)
    \label{eq:wrench_compensation_main}
\end{equation}
where \(\mathbf{w}^{raw}_t\) is the raw sensor reading, \(\mathbf{w}^{grav}_t\) denotes the pose-dependent gravity bias, and \(\mathcal{T}_{s\rightarrow ee}\) transforms the wrench from the sensor frame to the end-effector frame.

This makes the recorded wrench substantially cleaner and more suitable for downstream learning, thereby improving the physically grounded quality of the dataset. 
External-force acquisition in OmniUMI also benefits from the handheld interaction form itself. 
Because the operator directly feels part of the interaction wrench through proprioception and contact resistance, while the tool-side sensor records the same interaction quantitatively, the resulting data create a stronger correspondence between what the operator feels, what the system measures, and what the dataset stores. 
Thus, external force is not merely an additional sensor channel, but a grounded interaction variable that also contributes to human-aligned demonstration.

From a broader perspective, this design shows that force sensing in robot-free interfaces is not only a measurement problem, but also an embodiment problem: physically grounded and human-aligned wrench acquisition requires the sensing and actuation pathways to be co-designed.

\paragraph{Tactile Interaction and Visual Observations.}
In addition to internal grasping force and external interaction wrench, OmniUMI also acquires tactile interaction together with RGB, depth, and trajectory observations. Tactile sensing is not treated merely as an auxiliary modality, but as a human-aligned record of local contact deformation during manipulation.
These channels provide complementary information: tactile sensing captures local contact deformation and contact evolution, RGB and depth provide scene-level context and geometric structure, and trajectory provides motion reference. 
From a system perspective, tactile interaction complements internal and external force by preserving local contact detail, while RGB, depth, and trajectory provide supporting context for multimodal policy learning.
Their synchronized combination with internal and external force allows the collected dataset to retain both global context and local interaction detail within one robot-free multimodal interface.

Since the main contribution of this work does not lie in advancing these sensing channels individually, we keep this component lightweight and focus instead on their synchronized integration into the overall multimodal interface. 
In the current system, tactile images are first encoded into low-dimensional tactile embeddings, which are then combined with other low-dimensional observations before being fed to the policy. 
This choice preserves a compact policy interface while allowing the main methodological emphasis to remain on physically grounded multimodal data acquisition, human-aligned interaction acquisition, and deployment-compatible control.

\begin{figure*}[t]
    \centering
    \includegraphics[width=\textwidth]{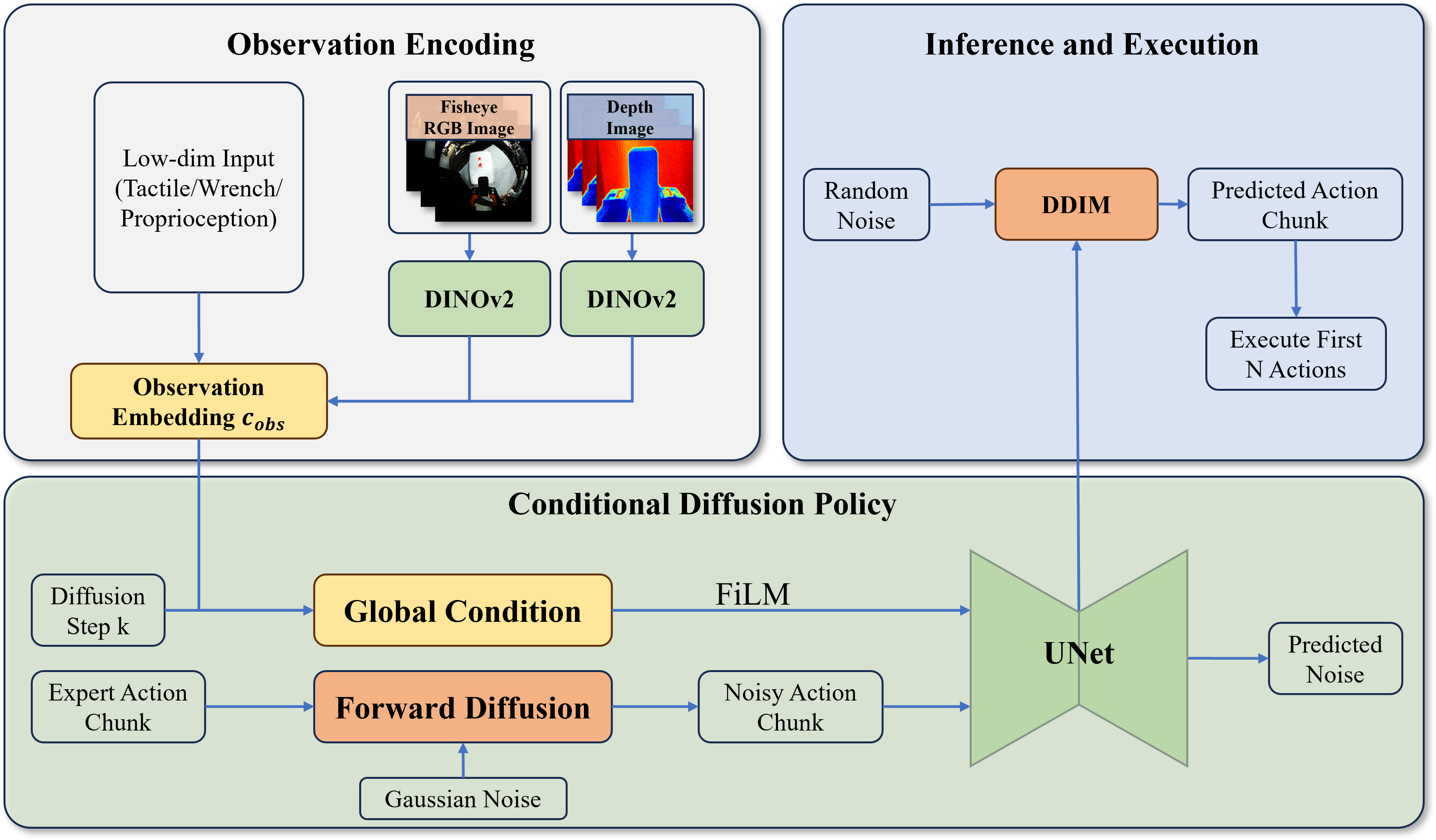}
    \caption{Overview of the multimodal diffusion-policy framework used in \textit{OmniUMI}. Low-dimensional observations (tactile, wrench, and proprioception) and visual inputs (fisheye RGB and depth) are encoded into an observation condition, which is combined with the diffusion-step embedding to form the global condition for the conditional U-Net. During training, the model predicts denoising trajectories from noisy action chunks, while during inference DDIM sampling produces action chunks that are executed in a receding-horizon manner.}
    \label{fig:omniumi_policy}
\end{figure*}

\subsection{Multimodal Policy Learning}

The final goal of data collection is not only to record multimodal interaction, but to make the resulting signals usable for downstream policy learning. 
At the same time, the policy interface should remain compatible with the controller used at deployment. 
Since the main contribution of this work does not lie in a new policy backbone, we adopt a policy architecture largely following UMI-FT and diffusion-policy-style visuomotor learning.

As illustrated in Fig.~\ref{fig:omniumi_policy}, the policy conditions the diffusion model on both multimodal observations and diffusion-step embeddings, and predicts action chunks that are executed in a receding-horizon manner. 
We first encode the tactile image into a compact tactile embedding
\begin{equation}
    \mathbf{z}^{tac}_t = \phi_{tac}\!\left(I^{tac}_t\right),
    \label{eq:tactile_embedding}
\end{equation}
where $\phi_{tac}$ denotes the tactile encoder.

The policy then maps synchronized multimodal observations to action predictions:
\begin{equation}
    \mathbf{a}_t = \pi_\theta(\mathbf{o}_t)
    \label{eq:policy_main}
\end{equation}
with
\begin{equation}
    \mathbf{o}_t
    =
    \left\{
        I^{rgb}_t,\,
        I^{depth}_t,\,
        \mathbf{z}^{tac}_t,\,
        \mathbf{p}_t,\,
        \mathbf{f}^{int}_t,\,
        \mathbf{f}^{ext}_t
    \right\}
    \label{eq:obs_main}
\end{equation}
where the observation includes synchronized visual inputs, tactile embeddings, kinematic signals, and force-related signals.

To preserve a clean interface between multimodal collection and controller-compatible execution, we adopt a compact action design:
\begin{equation}
    \mathbf{a}_t
    =
    \left[
        \Delta x_t,\,
        \Delta y_t,\,
        \Delta z_t,\,
        \mathbf{r}^{6D}_t,\,
        f^x_t,\,
        f^y_t,\,
        f^z_t,\,
        w^g_t
    \right]
    \label{eq:action_main}
\end{equation}
consisting of end-effector translation in \(xyz\), a 6D rotation representation, a predicted Cartesian interaction force \(\mathbf{f}_t = [f^x_t,f^y_t,f^z_t]^\top\), and a virtual gripper width.

This lightweight interface keeps the policy compatible with the downstream virtual-target-based controller without overemphasizing network novelty. 
During collection, the interface records how humans coordinate visual perception, local contact, and force regulation in a task-dependent manner. 
During learning, the policy absorbs these correlations as a mapping from multimodal observations to interaction-aware action variables. 
During deployment, these outputs are interpreted by the downstream controller in a physically meaningful way. 
In this sense, the policy interface is not merely a learning abstraction, but the central representation that links physically grounded sensing, human-aligned interaction, and controller-compatible robot execution.

\subsection{Impedance-Compatible Deployment}

The final component of OmniUMI is a deployment pathway that translates multimodal policy outputs into physically meaningful robot behavior~\cite{luo2025adaptive}. 
In contact-rich manipulation, the key challenge lies not only in predicting interaction-aware action variables, but also in executing them under the control constraints of practical robotic systems. 
To address this, OmniUMI adopts a unified deployment formulation based on virtual target pose and an impedance-compatible hybrid execution scheme.

\paragraph{Virtual Target Pose for Unified Interaction Representation.}
A central design goal of OmniUMI is to represent position- and force-related interaction intent within a single formulation, rather than treating motion control and force control as separate modes. 
In contact-rich manipulation, explicit mode switching often introduces discontinuity, instability, and cumbersome manual tuning. 
Instead, we translate both geometric action outputs and predicted interaction force into a unified virtual target pose.

We use a stiffness schedule modulated by the predicted force magnitude. 
Following the intuition of ACP-style compliance compilation, smaller predicted force corresponds to higher stiffness, while larger predicted force corresponds to lower stiffness. 
This allows the same Cartesian interaction representation to continuously interpolate between precision-oriented motion and compliant contact behavior. 
In the main text, we adopt a simplified diagonal form
\begin{equation}
    K_{p,t} = \mathrm{diag}(k_{x,t},\,k_{y,t},\,k_{z,t})
    \label{eq:kp_diag_main}
\end{equation}
where each stiffness term is scheduled from the predicted force magnitude. 
The resulting virtual target displacement and position are
\begin{equation}
    \Delta \mathbf{p}_t
    =
    K_{p,t}^{-1}\mathbf{f}_t,
    \quad
    \mathbf{p}^{vt}_t
    =
    \mathbf{p}^{ref}_t - \Delta \mathbf{p}_t
    \label{eq:delta_p_main_revised}
\end{equation}
with
\begin{equation}
    T^{vt}_t = \left(R^{ref}_t,\mathbf{p}^{vt}_t\right)
    \label{eq:vt_pose_main_revised}
\end{equation}

External-force-related interaction is thus expressed explicitly through Cartesian virtual target pose. 
For the gripper channel, the same principle is realized in a simpler form: the master-side gripper width is directly used as the target width of the slave gripper, and the tracking error under PD execution implicitly determines grasp regulation under contact. 
In this sense, internal-force-related interaction emerges from the grasp reference and the resulting gripper tracking behavior, rather than being explicitly controlled as a separate force variable.

\paragraph{Hybrid Impedance Execution.}
While the virtual target pose is defined in Cartesian space, practical robot systems are often executed through joint-space control interfaces, commonly implemented using impedance control. 
This results in a mismatch between interaction representation and execution, giving rise to two deployment gaps. 
The first is the gap between admittance-style virtual-target execution and impedance-based execution: although virtual-target formulations are effective in admittance-style settings, direct transfer to practical Cartesian impedance control is nontrivial. This is because lower achievable stiffness often leads to larger translational and rotational deviations under contact. 
The second is the gap between Cartesian interaction representation and joint-space execution: compliant behavior is defined in Cartesian space, but must ultimately be realized through a low-level control interface, typically implemented in joint space using impedance control.

To bridge these gaps, OmniUMI adopts a hybrid impedance execution strategy. 
Cartesian virtual target pose and stiffness are preserved as the interaction-level representation, and then translated into a controller-compatible form through the construction of operational-space gains, inverse kinematics, and joint-level impedance execution. 
We form the operational-space gain matrices as
\begin{equation}
    K_{x,t}=
    \begin{bmatrix}
        K_{p,t} & 0\\
        0 & K_R
    \end{bmatrix},
    \quad
    K_{xd,t}=
    \begin{bmatrix}
        D_{p,t} & 0\\
        0 & D_R
    \end{bmatrix}
    \label{eq:kx_main}
\end{equation}
and the corresponding effective joint-space gains as
\begin{equation}
    K^{(q)}_{p,t}=J^\top K_{x,t}J+K_q,
    \quad
    K^{(q)}_{d,t}=J^\top K_{xd,t}J+K_{qd}
    \label{eq:joint_gains_main}
\end{equation}
The control torque is given by
\begin{equation}
    \tau_t
    =
    K^{(q)}_{p,t}(q_{d,t}-q_t)
    +
    K^{(q)}_{d,t}(\dot q_{d,t}-\dot q_t)
    +
    C(q_t,\dot q_t)
    +
    g(q_t)
    \label{eq:tau_fb_main}
\end{equation}
In practice, the controller is implemented with inverse-dynamics compensation, incorporating Coriolis and gravity terms.
The desired joint states are obtained by damped least-squares inverse kinematics:
\begin{equation}
    \Delta q_t
    =
    J^\top
    (JJ^\top+\lambda^2I)^{-1}
    \,
    \xi\!\left(T^{vt}_t,T(q_t)\right),
    \quad
    q_{d,t}=q_t+\Delta q_t
    \label{eq:ik_main}
\end{equation}

This execution scheme preserves the Cartesian interaction representation of the policy output while deploying it through a joint-space impedance controller that is compatible with practical robotic systems. 
Rather than introducing a separate force-control branch, the controller interprets force-aware outputs as part of a unified execution target, allowing the robot to maintain precision where needed while remaining compliant under contact. 
A more complete treatment of hybrid impedance control, variable impedance scheduling, and passivity analysis is deferred to future work.

%% file: sec/4_experiment.tex
\section{Experiments}

We evaluate \textit{OmniUMI} from three complementary perspectives: 
(1) whether the proposed interface provides reliable \emph{physically grounded} sensing for contact-rich interaction, 
(2) whether the proposed \emph{human-aligned} interaction design improves the quality of collected demonstrations across internal grasp-force, external-force, and tactile alignment, and 
(3) whether the resulting multimodal data improve downstream policy learning on representative contact-rich manipulation tasks. 
Consistent with the central claim of this work, the downstream evaluations are organized around three key interaction capabilities enabled by the proposed framework: grasp-force-aware manipulation, wrench-informed surface interaction, and tactile-informed fine-grained control. 
Taken together, these experiments assess \textit{OmniUMI} not merely as a multimodal sensing platform, but as a unified robot-free framework for physically grounded robot learning via human-aligned multimodal interaction.

\subsection{Physically Grounded Sensing Verification}

Accurate sensing of interaction variables is a prerequisite for physically grounded robot learning. 
We first validate the reliability of external-wrench sensing in \textit{OmniUMI} by evaluating the gravity compensation procedure. 
Specifically, we measure 3D force signals at multiple static poses before and after compensation. 
The end-effector is moved through 10 distinct orientations spanning the workspace, and sensor readings are recorded for 1~s at each pose.

\begin{figure}[h]
    \centering
    \includegraphics[width=\linewidth]{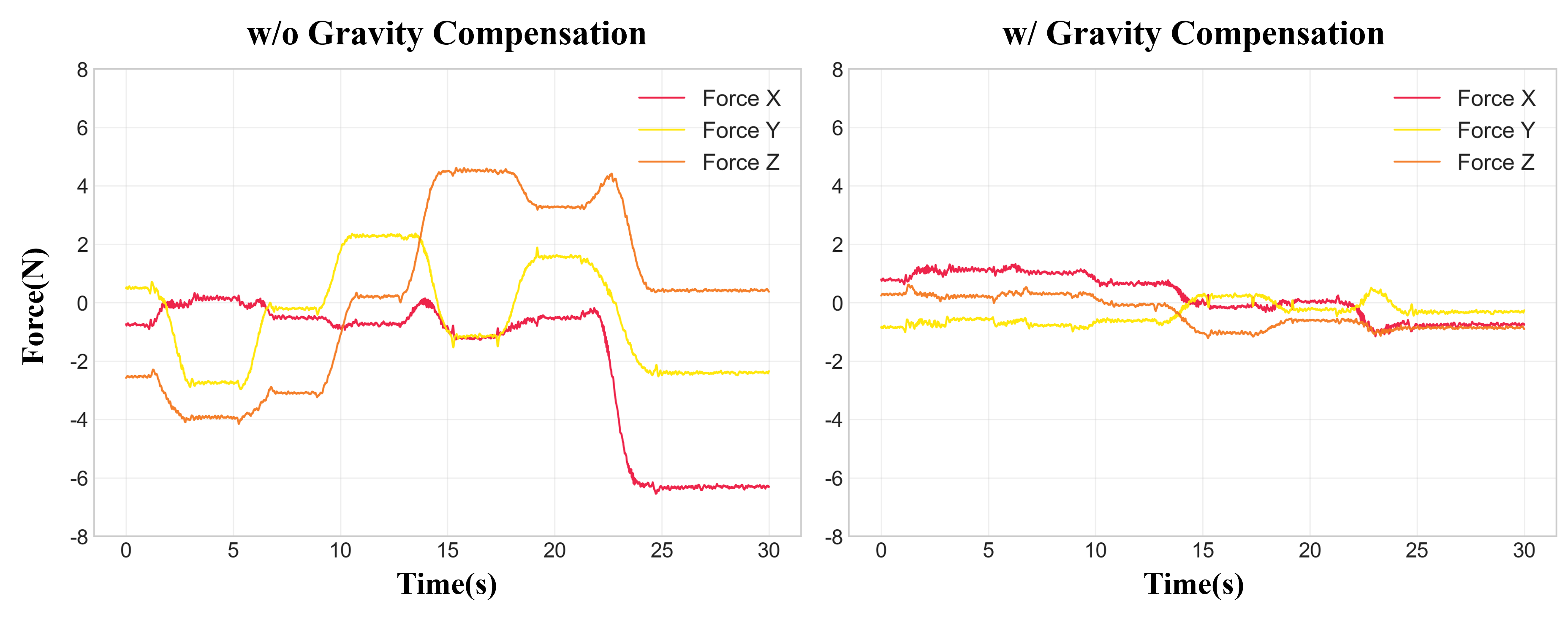}
    \caption{Representative force measurements before and after gravity compensation. 
    Left: raw force signals without gravity compensation, showing strong pose-dependent bias. 
    Right: compensated force signals after applying the calibrated gravity model and mass parameter identification, where the force measurements become significantly more concentrated around zero across different static poses.}
    \label{fig:gravity_comp}
\end{figure}

As shown in Fig.~\ref{fig:gravity_comp}, without compensation the raw force signals exhibit strong pose-dependent bias across all axes. 
Across different static poses, the force components vary by as much as 6.7~N. 
After applying the calibrated gravity model and mass parameter identification, the compensated force signals become significantly more concentrated around zero, with force variations substantially suppressed. 
These results verify that the proposed compensation model effectively removes gravity-induced and static offsets, providing a reliable near-zero baseline for subsequent dynamic interaction measurements.

This experiment confirms that \textit{OmniUMI} can stably acquire one of the central physically grounded interaction variables used throughout this work, namely external interaction wrench. 
It also establishes the sensing reliability required for the downstream learning experiments that follow.

\subsection{Human-Aligned Demonstration Quality}

We next evaluate whether \textit{OmniUMI} improves the quality of collected demonstrations through its human-aligned interaction design. 
Rather than using different tasks for different modalities, we use a single contact-rich manipulation task to examine three forms of alignment in a unified manner: internal grasp-force alignment, external-force alignment, and tactile alignment. 
The task requires the operator to pick up a whiteboard eraser, erase red marker traces on the whiteboard, and then place the eraser back. 
Across all three evaluations, we compare the same three demonstration settings: direct human demonstration, teleoperation without bilateral force feedback, and \textit{OmniUMI} with bilateral gripper feedback and a handheld embodiment. 
For each setting, we analyze one representative trajectory from the same task and examine whether the recorded interaction signals are closer to natural human behavior across multiple physical channels, including internal grasp force, external contact force, and tactile deformation.

\paragraph{Internal grasp-force alignment.}
We first evaluate human alignment from the perspective of internal grasp force. 
Although the overall task includes grasping, wiping, and object placement, the internal grasp-force signal is most directly related to how the operator picks up, holds, and releases the eraser throughout the task. 
We therefore focus on the temporal evolution of the recorded grasp-force trajectory as the primary indicator of internal-force alignment.

\begin{figure}[h]
    \centering
    \includegraphics[width=\linewidth]{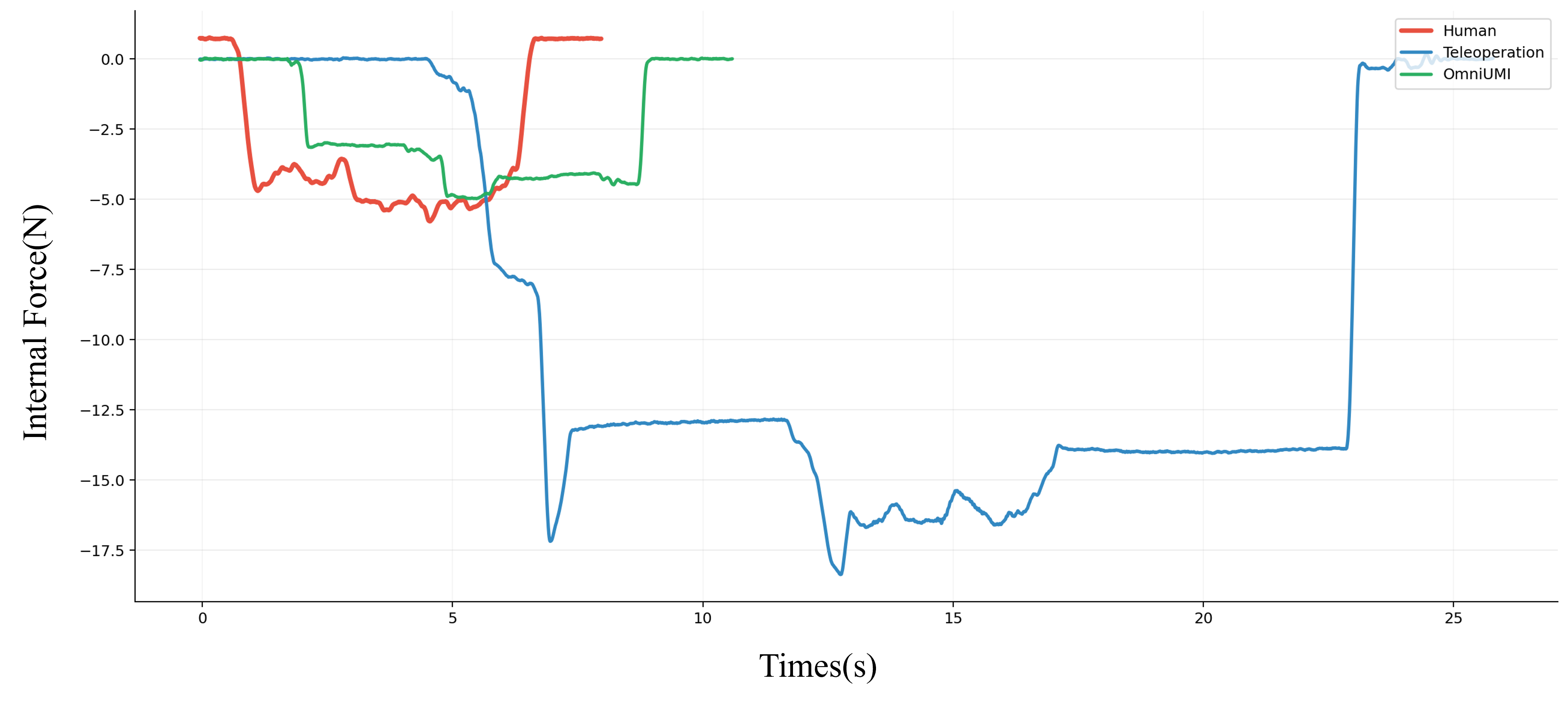}
    \caption{Representative internal grasp-force trajectories recorded during the whiteboard eraser pick--wipe--place task under three demonstration settings: direct human demonstration, teleoperation without bilateral force feedback, and \textit{OmniUMI} with bilateral gripper feedback and a handheld embodiment. 
    \textit{OmniUMI} produces grasp-force trajectories that are substantially closer to direct human demonstrations in both overall magnitude and temporal modulation, whereas teleoperation without bilateral force feedback leads to less stable and less human-like force regulation.}
    \label{fig:grasp_force_comparison}
\end{figure}

As shown in Fig.~\ref{fig:grasp_force_comparison}, the internal grasp-force trajectory collected with \textit{OmniUMI} more closely matches that of direct human demonstration, both in overall magnitude and in temporal variation. 
In contrast, the teleoperation baseline produces a less stable force trajectory and larger deviations from the human reference, indicating poorer regulation of grasp interaction during demonstration.

This result suggests that internal grasp-force alignment cannot be achieved reliably through visual observation alone. 
Without bilateral force feedback, teleoperation users must rely primarily on visual observation and therefore have difficulty reproducing the timing and magnitude of human-like grasp-force modulation. 
By contrast, \textit{OmniUMI} combines bilateral gripper feedback with a handheld embodiment, enabling the operator to regulate grasp interaction in a more natural and embodied manner and resulting in smoother and more human-like internal-force trajectories.

\paragraph{External-force alignment.}
We next evaluate human alignment from the perspective of external contact force using the same task. 
Among the different phases of the trajectory, the wiping stage is the most contact-critical, because successful erasing depends on maintaining stable contact with the whiteboard surface while removing the red marker traces. 
Accordingly, we focus on the recorded normal contact force along the $z$ axis during wiping, i.e., the $F_z$ profile, as the primary indicator of external-force alignment.

\begin{figure}[h]
    \centering
    \includegraphics[width=\linewidth]{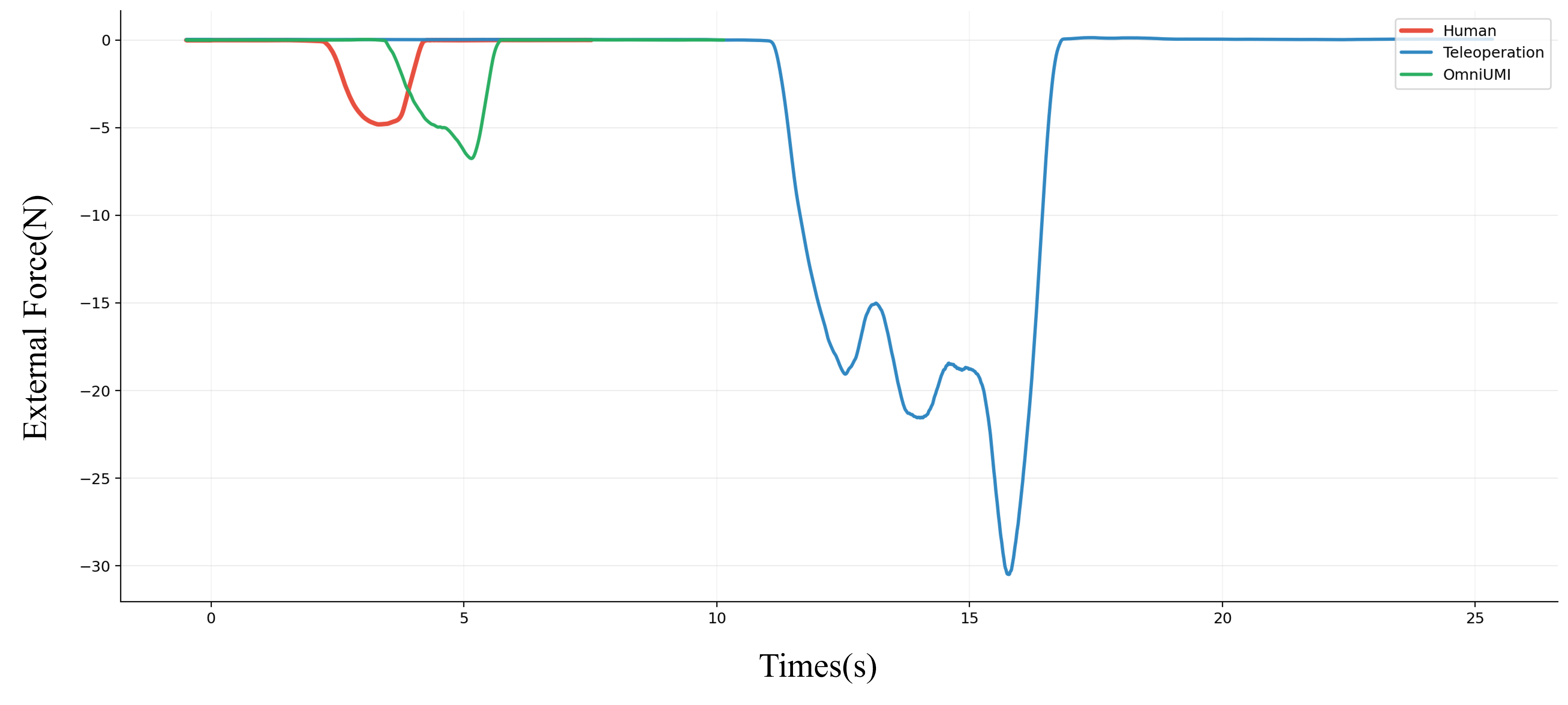}
    \caption{Representative $F_z$ profiles recorded during the wiping phase of the whiteboard eraser pick--wipe--place task under three demonstration settings: direct human demonstration, teleoperation without bilateral force feedback, and \textit{OmniUMI} with bilateral gripper feedback and a handheld embodiment. 
    Since effective whiteboard cleaning depends critically on stable normal contact during erasing, the $z$-axis force provides the most informative comparison. 
    \textit{OmniUMI} produces force trajectories that are much closer to direct human demonstrations, whereas teleoperation exhibits more severe force oscillations during wiping.}
    \label{fig:teleop_comparison}
\end{figure}

As shown in Fig.~\ref{fig:teleop_comparison}, the $F_z$ trajectory collected with \textit{OmniUMI} during the wiping phase is much closer to that of direct human demonstration. 
In contrast, the teleoperation baseline produces highly oscillatory force signals, indicating unstable contact regulation in the most interaction-sensitive stage of the task.

These observations suggest that, without bilateral force feedback, teleoperation users rely mainly on visual cues and tend to overcompensate at contact, which leads to unstable force modulation. 
By contrast, \textit{OmniUMI} combines bilateral gripper feedback with a handheld structure that preserves more natural embodied contact regulation during wiping, thereby producing external-force trajectories that better match direct human demonstrations.

\paragraph{Tactile alignment.}
We further evaluate human alignment from the perspective of tactile interaction, again using the same whiteboard eraser pick--wipe--place task. 
To quantify tactile interaction, we use the marker-motion magnitude on the tactile image as the evaluation metric. 
Specifically, let $\Delta \mathbf{m}_t \in \mathbb{R}^{126}$ denote the marker offset vector at time step $t$, obtained by concatenating the 2D displacements of all tactile markers. 
We then compute its $\ell_2$ norm, $\|\Delta \mathbf{m}_t\|_2$, which measures the overall deformation magnitude of the tactile image over time. 
Because this quantity directly reflects the strength and temporal variation of local contact-induced deformation, it provides an informative proxy for tactile alignment.

\begin{figure}[h]
    \centering
    \includegraphics[width=\linewidth]{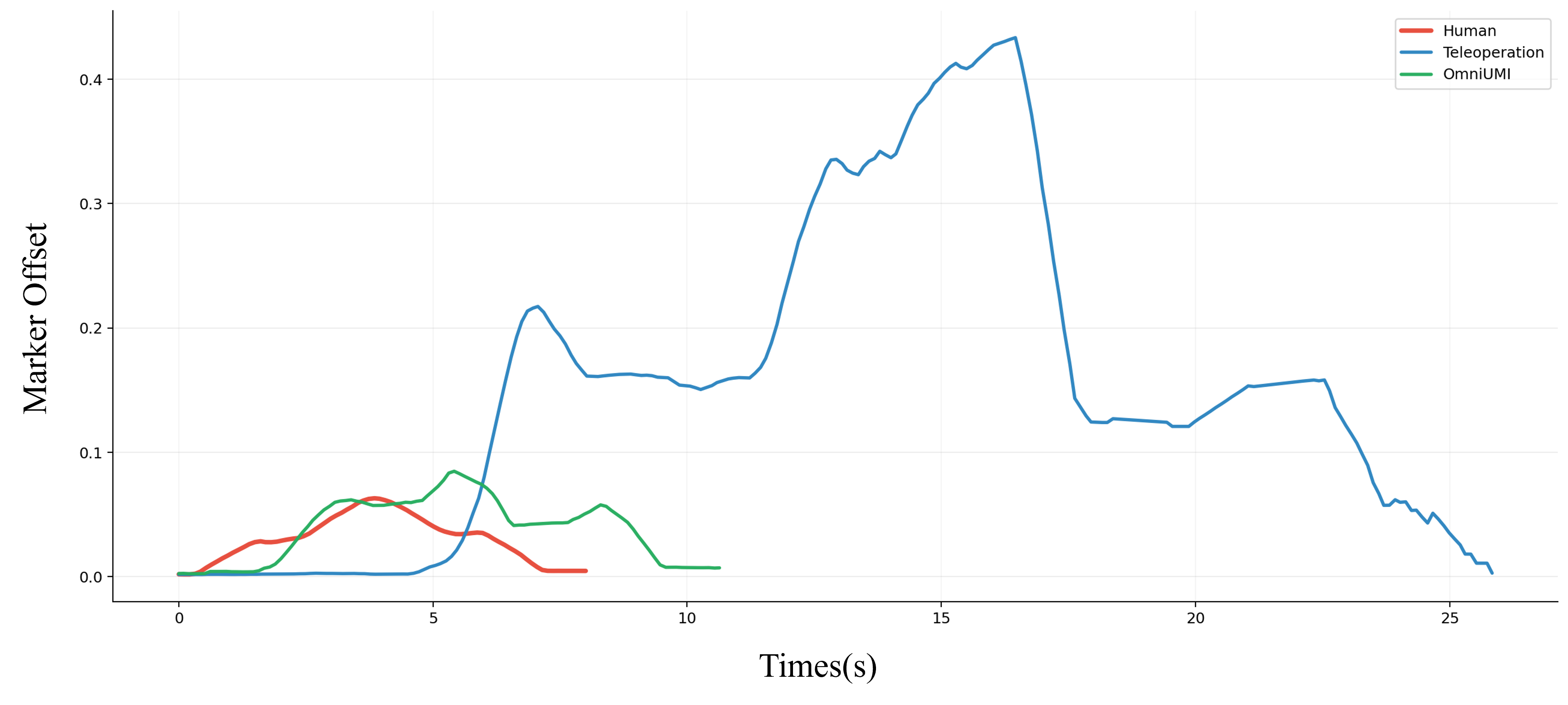}
    \caption{Representative tactile interaction trajectories recorded during the whiteboard eraser pick--wipe--place task under three demonstration settings: direct human demonstration, teleoperation without bilateral force feedback, and \textit{OmniUMI} with bilateral gripper feedback and a handheld embodiment, measured by the $\ell_2$ norm of the 126-dimensional marker offset vector over time. 
    \textit{OmniUMI} produces tactile trajectories that are substantially closer to direct human demonstrations in both amplitude and temporal variation, whereas teleoperation yields noisier and less consistent tactile patterns.}
    \label{fig:tactile_alignment}
\end{figure}

As shown in Fig.~\ref{fig:tactile_alignment}, the tactile trajectory collected with \textit{OmniUMI} more closely matches that of direct human demonstration, both in overall deformation magnitude and in temporal evolution. 
In contrast, the teleoperation baseline produces tactile signals that are more irregular and less consistent with the human reference, indicating reduced fidelity in contact perception and modulation during demonstration.

\begin{figure*}[t]
    \centering
    \includegraphics[width=\textwidth]{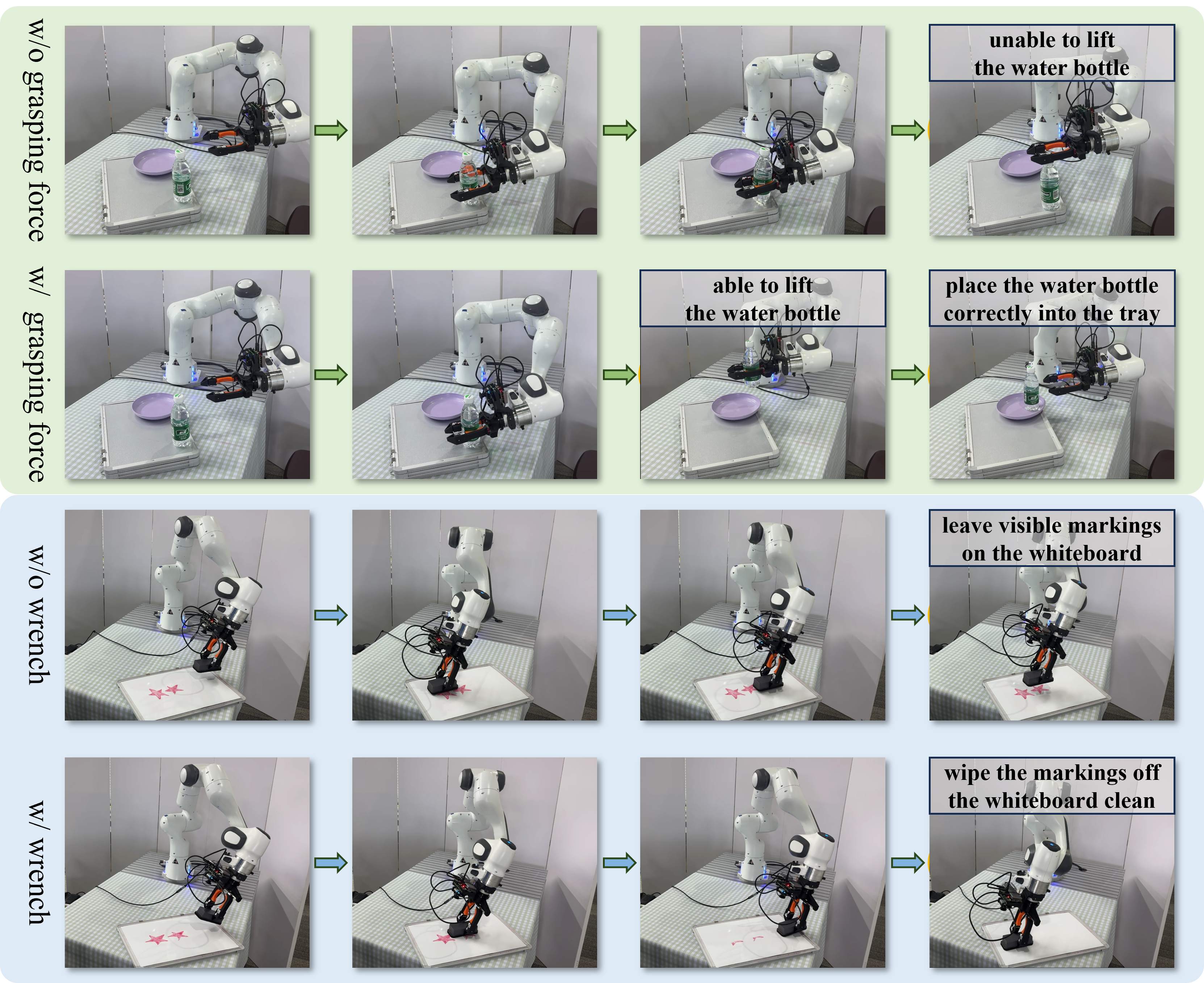}
    \caption{Qualitative results on force-related contact-rich manipulation tasks. 
    \textbf{Top:} force-sensitive manipulation with a full unopened bottle of mineral water. 
    Without grasp-force awareness, the policy fails to lift the bottle; with the proposed force-aware policy, the robot successfully lifts and places the bottle correctly into the tray. 
    \textbf{Bottom:} interactive wrench-based whiteboard erasing. 
    Without wrench-aware control, visible markings remain on the whiteboard; with the proposed wrench-informed policy, the robot successfully wipes the markings off the whiteboard clean.}
    \label{fig:force_wrench_qualitative}
\end{figure*}

These results suggest that the human-aligned embodiment of \textit{OmniUMI} improves not only internal grasp-force modulation and external-force regulation, but also tactile interaction acquisition. 
Without bilateral force feedback, teleoperation operators rely primarily on visual observation and therefore have greater difficulty reproducing the fine-grained temporal structure of human tactile interaction. 
By contrast, \textit{OmniUMI} combines bilateral gripper feedback with a handheld embodiment, enabling more natural and physically grounded interaction with the environment and producing tactile trajectories whose temporal structure better matches human contact behavior.

Taken together, these results support the claim that human alignment should be evaluated across multiple physical interaction channels within the same contact-rich manipulation task rather than through vision alone. 
Across internal grasp force, external contact force, and tactile deformation, \textit{OmniUMI} consistently produces demonstration trajectories that are closer to direct human behavior than conventional teleoperation, validating the effectiveness of its human-aligned multimodal interaction design.

\subsection{Downstream Contact-rich Manipulation}

We finally evaluate whether the physically grounded and human-aligned multimodal data collected by \textit{OmniUMI} improve downstream policy learning in representative contact-rich manipulation settings. 
Following the central formulation of this work, the tasks are organized around three key interaction capabilities: grasp-force-aware manipulation, wrench-informed surface interaction, and tactile-informed fine-grained control.

\begin{figure*}[t]
    \centering
    \includegraphics[width=\textwidth]{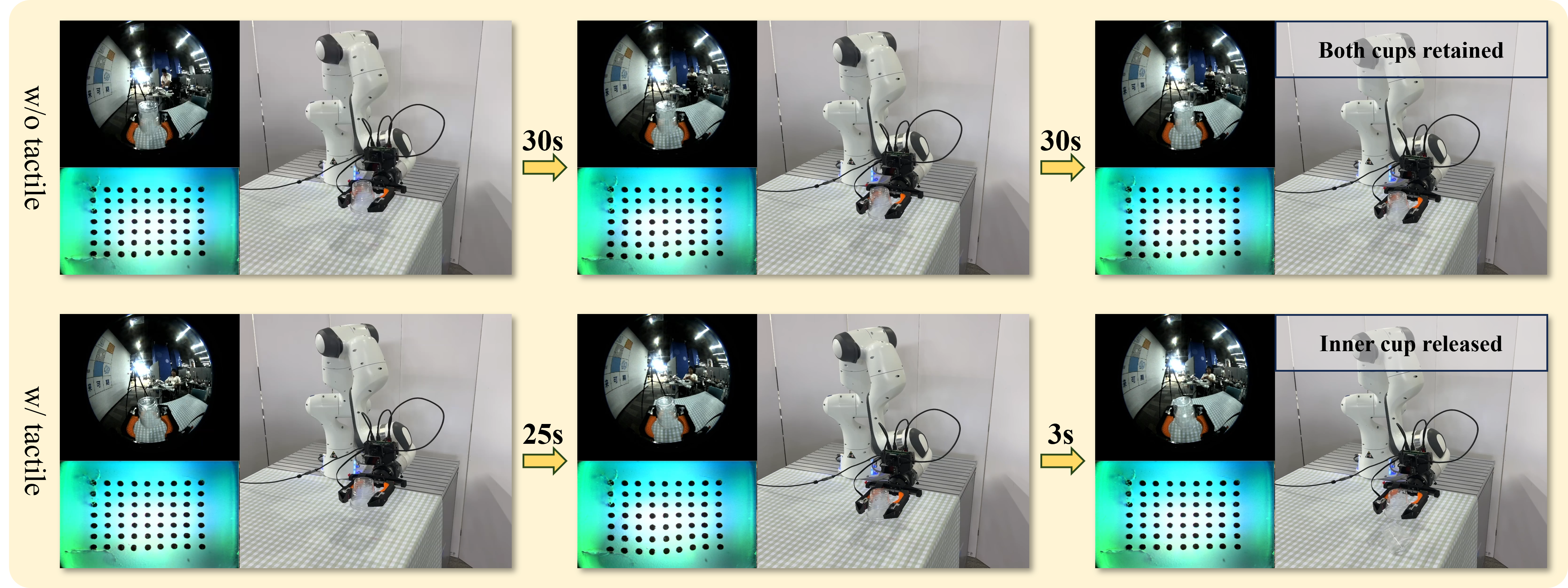}
    \caption{Qualitative results on tactile-informed selective release. 
    The robot initially grasps two nested transparent cups under minimal gripping force. 
    \textbf{Top:} failure case without sufficient tactile discrimination, where both cups are retained. 
    \textbf{Bottom:} success case with tactile-informed control, where the inner cup is released while the outer cup remains grasped. 
    The accompanying fisheye and tactile observations illustrate the local contact cues used for fine-grained release control.}
    \label{fig:tactile_release_qualitative}
\end{figure*}

\paragraph{Grasp-Force-Aware Manipulation.}
We first evaluate downstream learning in an internal-force-dominated manipulation setting. 
Specifically, we perform a pick-and-place task using a relatively heavy object, namely a full unopened bottle of mineral water. 
This task requires sufficiently large and well-regulated internal grasping force to securely lift, transport, and place the object without slippage, making it an appropriate benchmark for load-sensitive grasp control.

Policies are trained using multimodal data collected with \textit{OmniUMI} and are evaluated based on their ability to apply adequate grasping force while maintaining stable task execution. 
In this setting, insufficient force typically leads to grasp failure or slippage during lifting and transport, whereas excessive force is unnecessary and indicates poor regulation.

Compared with policies trained without internal-force input, the force-aware variant achieves substantially higher success rates in transporting the full water bottle. 
Figure~\ref{fig:force_wrench_qualitative} (top) provides qualitative comparisons. 
Without grasp-force awareness, the policy fails to lift the water bottle, whereas the proposed force-aware policy successfully lifts and places it into the tray. 
These results confirm that physically grounded internal-force signals provide effective supervision for learning stable grasp regulation under nontrivial load conditions.

\begin{table}[h]
\centering
\caption{Comparison on the force-sensitive pick-and-place task using a full unopened bottle of mineral water. Higher success rate and lower slippage indicate better internal grasp-force regulation.}
\label{tab:force_sensitive_results}
\resizebox{\linewidth}{!}{
\begin{tabular}{lcc}
\toprule
Method & Success Rate (\%) $\uparrow$ & Slippage Rate (\%) $\downarrow$ \\
\midrule
w/o Grasping Force & 0.0 & 100.0 \\
w/ Grasping Force (Ours) & \textbf{100.0} & \textbf{0.0} \\
\bottomrule
\end{tabular}
}
\end{table}

\paragraph{Wrench-Informed Surface Interaction.}
We next evaluate downstream learning in an external-wrench-dominated interaction setting. 
To this end, we design a whiteboard erasing task in which the end-effector must remove red drawings from a flat whiteboard surface while maintaining continuous sliding contact. 
Compared with simple surface tracing, this task imposes a more realistic and interaction-intensive requirement, since successful erasing depends not only on trajectory tracking but also on maintaining appropriate external contact force throughout the motion.

The collected data exhibit a clear correlation between tangential wrench and end-effector velocity, capturing the stick--slip dynamics that naturally arise during real contact transitions on the board surface.

Policies trained on \textit{OmniUMI} erasing data successfully reproduce the contact trajectory while maintaining a normal force above 7~N along the $z$ axis, thereby enabling reliable removal of the red markings. 
In contrast, baselines trained without wrench input exhibit noticeable high-frequency oscillations due to unstable contact estimation, leading to less stable erasing behavior and degraded task performance. 
Figure~\ref{fig:force_wrench_qualitative} (bottom) further shows that wrench-informed control produces stable erasing behavior, while the policy without wrench input leaves visible markings on the whiteboard. 
These results demonstrate that \textit{OmniUMI} can both record and leverage dynamic external interaction wrench for learning smooth and compliant surface-contact manipulation skills.

\begin{table}[h]
\centering
\caption{Comparison on the interactive wrench-based whiteboard erasing task. Two success criteria are reported based on the remaining red-marking area after erasing.}
\label{tab:wrench_erasing_results}
\resizebox{\linewidth}{!}{
\begin{tabular}{lcc}
\toprule
Method & Success Rate@5\% Residual (\%) $\uparrow$ & Success Rate@50\% Residual (\%) $\uparrow$ \\
\midrule
w/o Wrench & 0.0 & 60.0 \\
w/ Wrench (Ours) & \textbf{100.0} & \textbf{100.0} \\
\bottomrule
\end{tabular}
}
\end{table}

\paragraph{Tactile-Informed Fine-Grained Control.}
Finally, we evaluate tactile-informed fine-grained control through a nested-cup selective-release task. 
The robot initially grasps two transparent plastic cups placed upside down and nested together. 
During execution, the policy is required to gradually loosen the gripper until the inner cup is released and drops, while the outer cup remains stably grasped. 
This task is intentionally designed to operate under very small gripping forces, making it a suitable benchmark for evaluating whether tactile sensing can support subtle contact-state discrimination and precise release control beyond what vision alone can provide.

Unlike conventional grasping tasks, the challenge here is not to maximize grasp stability, but to regulate contact at the boundary between retention and release. 
Because the two cups are transparent, geometrically similar, and tightly nested, visual observations provide only limited information about their local contact state and relative motion. 
Successful execution instead depends on detecting subtle tactile changes as the inner cup begins to separate, while still maintaining sufficient control over the outer cup.

During deployment, the policy adjusts the gripper command primarily based on tactile input to achieve selective release under minimal gripping force. 
Quantitative results (Table~\ref{tab:tactile_results}) show that incorporating tactile embeddings substantially improves selective-release success while reducing failure modes such as retaining both cups. 
The tactile-informed policy learns to detect subtle changes in local contact conditions and uses this information to modulate release timing more precisely during execution. 

As illustrated in Fig.~\ref{fig:tactile_release_qualitative}, tactile information enables reliable discrimination between retention and release: without sufficient tactile information both cups are retained, whereas the tactile-informed policy successfully releases the inner cup while keeping the outer cup grasped. 
Qualitative observations further show that tactile input leads to more stable and reliable separation behavior in this low-force manipulation regime, validating the effectiveness of the tactile representations used in our framework.

\begin{table}[h]
\centering
\caption{Comparison on the tactile-informed nested-cup selective-release task. Tactile input improves selective release under minimal gripping force.}
\label{tab:tactile_results}
\resizebox{\linewidth}{!}{
\begin{tabular}{lcc}
\toprule
Method & Selective Release Success Rate (\%) $\uparrow$ & Both Cups Retained Rate (\%) $\downarrow$ \\
\midrule
w/o Tactile & 20.0 & 80.0 \\
w/ Tactile (Ours) & \textbf{100.0} & \textbf{0.0} \\
\bottomrule
\end{tabular}
}
\end{table}

\subsection{Summary}

Across all experiments, \textit{OmniUMI} demonstrates reliable physically grounded sensing, improved human-aligned demonstration quality, and strong downstream learning performance. 
Accurate gravity compensation provides reliable external-wrench baselines, while the unified sensing and embodiment design enables high-fidelity multimodal data collection at scale. 
Compared with conventional teleoperation without bilateral force feedback, \textit{OmniUMI} yields interaction signals that are consistently closer to direct human behavior across internal grasp force, external interaction wrench, and tactile deformation.
Downstream evaluations further show that the proposed framework effectively supports grasp-force-aware manipulation, wrench-informed surface interaction, and tactile-informed fine-grained control. 
Taken together, these results establish \textit{OmniUMI} as a practical foundation for physically grounded robot learning via human-aligned multimodal interaction beyond vision-only representations.

%% file: sec/5_conclusion.tex
\section{Conclusion}

We presented \textit{OmniUMI}, a unified framework for \emph{physically grounded robot learning via human-aligned multimodal interaction}. 
Unlike prior UMI-like systems that primarily focus on visuomotor recording, \textit{OmniUMI} moves toward interaction acquisition by jointly capturing tactile sensing, internal grasping force, external interaction wrench, and visual-motion observations within a single handheld platform, while organizing their acquisition around a human-aligned demonstration interface that supports natural perception and regulation of multimodal physical interaction.

The proposed framework is built around three tightly coupled components: a unified multimodal handheld interface for physically grounded data acquisition, a human-aligned interaction acquisition pipeline grounded in bilateral gripper feedback and the handheld embodiment, and a multimodal learning and impedance-compatible deployment framework. 
Together, these components establish a consistent pipeline from robot-free multimodal human interaction to contact-rich robot execution.

At the system and data-acquisition levels, \textit{OmniUMI} supports practical multimodal integration, collection--deployment consistency, and physically grounded multimodal sensing within a shared embodiment. 
At the interaction-acquisition level, it improves human alignment during demonstration by enabling the operator to naturally perceive and regulate internal grasping force, external interaction wrench, and tactile interaction through bilateral gripper feedback and the handheld embodiment. 
At the learning and deployment levels, \textit{OmniUMI} connects physically grounded and human-aligned multimodal data to downstream policy learning by extending diffusion policy with visual, tactile, and force-related observations, and translates the resulting policy outputs into controller-compatible execution through impedance-based deployment.

Experiments show that \textit{OmniUMI} improves sensing reliability, human-aligned demonstration quality, and downstream manipulation performance. 
In particular, the proposed framework supports three key interaction capabilities: grasp-force-aware manipulation, wrench-informed surface interaction, and tactile-informed fine-grained control, validated respectively on force-sensitive pick-and-place, interactive surface erasing, and tactile-informed selective release. 
These results demonstrate that physically grounded multimodal interaction data can be reliably acquired, naturally perceived and regulated by human operators across internal grasp force, external contact force, and tactile interaction, and effectively leveraged for downstream contact-rich manipulation within a unified robot-free framework.

At the same time, several limitations remain. 
Although the current results show improved signal-level alignment with human demonstration, a fuller causal account of how such alignment contributes to downstream policy performance is still lacking. 
In particular, while our experiments indicate that bilateral gripper feedback and the handheld embodiment improve the quality of collected interaction signals across internal grasp force, external contact force, and tactile interaction, and benefit subsequent learning, the precise relationship between human-aligned interaction acquisition, data quality, and final task performance has not yet been isolated in a systematic manner. 
More broadly, the current study is evaluated on a limited set of contact-rich tasks and a single system instantiation, and a more comprehensive analysis across operators, embodiments, and task families remains necessary.

These limitations point to several directions for future work. 
An important next step is to study human alignment more explicitly, for example through controlled ablations, user studies, and finer-grained analysis of how bilateral feedback and handheld embodiment jointly shape internal-force, external-force, and tactile interaction during demonstration and downstream learning. 
It is also important to extend the framework to a broader range of embodiments and interaction settings, including more dexterous manipulation, longer-horizon tasks, and more diverse contact conditions. 
On the learning and control side, future work may further strengthen the coupling between multimodal policy learning and controller-compatible execution, including richer interaction representations and more principled impedance-aware deployment strategies.

Taken together, our findings suggest that progress in contact-rich robot learning depends not only on stronger policy models, but also on better interfaces for acquiring, aligning, and deploying physically meaningful interaction data. 
By combining physically grounded multimodal data acquisition with human-aligned interaction, \textit{OmniUMI} offers a practical step beyond visuomotor recording toward scalable robot-free learning for contact-rich manipulation.